\documentclass[conference]{IEEEtran}
\IEEEoverridecommandlockouts
\usepackage{cite}
\usepackage{svg}
\usepackage{bm}
\usepackage{amsmath,amssymb,amsfonts}
\usepackage{adjustbox}
\usepackage{algorithmic}
\usepackage{graphicx}
\usepackage{subcaption}
\usepackage{textcomp}
\usepackage{bm}
\usepackage{url}
\usepackage{optidef}
\usepackage{xcolor}
\usepackage{caption}
\usepackage{subcaption}
\usepackage{multirow}
\usepackage{diagbox}

\DeclareMathOperator*{\argmin}{arg\,min}
\def\BibTeX{{\rm B\kern-.05em{\sc i\kern-.025em b}\kern-.08em
    T\kern-.1667em\lower.7ex\hbox{E}\kern-.125emX}}
\begin{document}

\title{Smooth path planning with safety margins using Piece-Wise Bezier curves}

\author{
    \IEEEauthorblockN{
    Iancu Andrei\IEEEauthorrefmark{1},
    Marius Kloetzer\IEEEauthorrefmark{1},
    Cristian Mahulea\IEEEauthorrefmark{2},
    Catalin Dosoftei\IEEEauthorrefmark{1}}
    \IEEEauthorblockA{\IEEEauthorrefmark{1}Faculty of Automatic Control and Control Engineering, "Technical University of Iasi, Romania\\
    \{andrei-iulian.iancu, marius.kloetzer, constantin-catalin.dosoftei\}@academic.tuiasi.ro.}
    \IEEEauthorblockA{\IEEEauthorrefmark{2}Arag\'on Institute of Engineering Research (I3A), University of Zaragoza, Spain\\ cmahulea@unizar.es.}
    \thanks{This work was supported in part by grants PID2021-125514NB-I00 and PID2021-125514NB-I00 funded by MCIN/AEI/10.13039/501100011033 and by the European Union NextGenerationEU/PRTR, and by Grant N62909-24-1-2081 funded by the Office of Naval Research Global, USA.}
}
\maketitle

\begin{abstract}

In this paper, we propose a computationally efficient quadratic programming (QP) approach for generating smooth, $C^1$ continuous paths for mobile robots using piece-wise quadratic Bezier (PWB) curves. Our method explicitly incorporates safety margins within a structured optimization framework, balancing trajectory smoothness and robustness with manageable numerical complexity suitable for real-time and embedded applications. Comparative simulations demonstrate clear advantages over traditional piece-wise linear (PWL) path planning methods, showing reduced trajectory deviations, enhanced robustness, and improved overall path quality. These benefits are validated through simulations using a Pure-Pursuit controller in representative scenarios, highlighting the practical effectiveness and scalability of our approach for safe navigation.

\end{abstract}


\section{Introduction}

Path planning for mobile robots is evolving from classical deterministic methods to more flexible probabilistic and intelligent approaches \cite{tang2025path}.
While classical methods offer predictable performance but limited robustness to uncertainty, probabilistic and learning-based strategies better handle dynamic, high-dimensional environments at the cost of non-deterministic, unpredictable outcomes.
Both classical and modern methods often overlook robustness and safety, commonly producing piecewise linear (PWL) paths that may be unattainable due to kinematic constraints.

To address safety explicitly, Model Predictive Control (MPC) and Control Barrier Functions (CBF) approaches have been introduced \cite{hwang2024safe}, enforcing constraints through iterative optimization. However, their scalability is limited, as computational complexity increases with system dynamics and environmental detail, posing challenges for real-time, embedded deployment.

A promising compromise between complexity, computational efficiency, and safety is provided by smooth path planning techniques, notably using Bezier or B-spline curves. Recent studies have demonstrated that employing smooth trajectories can significantly enhance control performance and robustness, while maintaining computational tractability. Some works utilize smooth curves to construct reference trajectories better suited to system dynamics \cite{lelidis2024planning}, while others employ nonlinear gradient-based optimization techniques considering curvature limits \cite{pool2023optimizing, wang2023autonomous}. Additional studies propose iterative algorithms for creating safe corridors, thereby implicitly enforcing safety margins \cite{wang2023autonomous, nguyen2023navigation}, or incorporate advanced temporal specifications such as Linear Temporal Logic (LTL) to ensure trajectory correctness in complex scenarios \cite{kurtz2023temporal}.

In this paper, we propose an efficient quadratic programming (QP) approach for generating safe and smooth $C^1$ continuous paths using piece-wise quadratic Bezier (PWB) curves. Our formulation explicitly incorporates safety margin constraints through the construction of safe polytopes, achieving robustness without significantly increasing numerical complexity.

The main contributions of this approach are:
\begin{itemize}
    \item consideration for robotic path trajectory performance through the use of Bezier curve properties with improved scalability through the use of quadratic programming;
    \item ability to plan for arbitrary robotic dimensions without recomputing the environment decomposition using the safe polytope construction.
\end{itemize}

\section{Methodology and problem description}

\textbf{Problem statement}. We consider a bounded two-dimensional (2D) robotic workspace populated with convex polygonal obstacles and defined workspace boundaries. The primary objective is to generate a trajectory from an initial pose to a goal pose that simultaneously minimizes total path length and absolute curvature, two essential criteria for efficient and safe robotic navigation. Minimizing path length directly contributes to shorter execution times, while curvature minimization reduces mechanical stress, prevents slippage, and ensures smooth, robust operation. The proposed solution must maintain computational efficiency and scalability, suitable for real-time implementation.

Given the workspace, a cell decomposition approach is used to partition the free space into convex polytopes, which forms a connectivity graph representing feasible navigation areas. A sequence of polytopes $\mathcal{P}^0, \ldots, \mathcal{P}^M$ is obtained using a standard graph-search algorithm, where the initial and final robot poses reside within polytopes $\mathcal{P}^0$ and $\mathcal{P}^M$.

\begin{figure}[h!]
    \centering
    \includegraphics[width=0.7\linewidth]{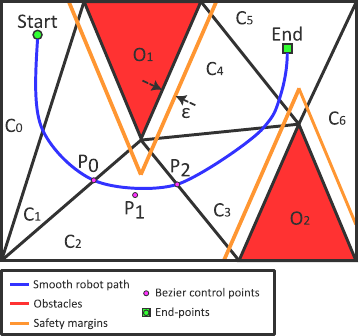}
    \caption{Safe and smooth robotic motion example.}
    \label{fig:workspace}
\end{figure}

To ensure obstacle avoidance and robustness, each polytope is modified into a ``safe polytope,'' effectively embedding a predefined safety margin. Subsequently, each polytope has an associated quadratic Bezier curve segment $B_0(t), \ldots, B_M(t)$, characterized by control points $P_j^i$, $j = 0, \ldots, n$, $i = 0, \ldots, M$. These control points serve as decision variables.
Fig.~\ref{fig:workspace} illustrates an example scenario using triangular decomposition.

\textbf{Convex Polytopes}. A convex polytope $\mathcal{P} \in \mathbb{R}^{N}$  is defined as the bounded intersection of a finite set of half-spaces, described mathematically by linear inequalities as: 
\begin{equation}
    \mathcal{P} = \{x\in \mathbb{R}^N | H \cdot x \le \overline{b}\},
\end{equation}
where $H \in \mathbb{R}^{K \times N}$ and $\overline{b}\in \mathbb{R}^K$.  Here, $K$ is the number of linear constraints (typically equal to the number of polytope edges), and $N$ is the workspace dimension, in our case $N =2$.

\textbf{Bezier curves} are polynomial curves employed in curve fitting due to their desirable geometric and numerical properties, especially for creating smooth trajectories. A Bezier curve of order $n$ is defined by:
\begin{equation}
    B(t) = \sum\limits_{i=0}^{n} b_i^n(t)P_i \quad t\in[0,1],
    \label{eq:bezier}
\end{equation}
 where $P_i \in \mathbb{R}^2$ are the control points defining the curve geometry, and $b_i^n(t)$ represents Bernstein polynomials:
\begin{equation}
    b_i^n(t) = {n \choose i} (1-t)^{n-i}t^i, \quad i \in {0,...,n}.
    \label{eq:bernstein}
\end{equation}

Key properties of Bezier curves used for our approach are:

\begin{itemize}
\item The entire Bezier curve segment lies within the convex hull formed by its control points.
\item Smoothness constraints $C^1$ (continuity up to first derivative), are implemented by enforcing linear constraints on control points.
\end{itemize}

For our specific formulation, quadratic (second-order) Bezier curves are chosen due to their optimal balance between computational simplicity and sufficient smoothness and curvature properties. A quadratic Bezier curve with control points $P_0, P_1, P_2$ is explicitly given by:
\begin{equation}
    \label{eq:quad_bezier}
    B(t) = (1-t)^2 P_0 + 2 (1-t) t P_1 + t^2 P_2, t \in [0,1].
\end{equation}
\noindent This choice provides flexibility and smoothness for robotic paths while maintaining efficient optimization complexity.

\textbf{Optimization problem formulation} for safe, smooth trajectory planning as is defined as follows:
\begin{equation}\label{eq:prob_ideal}
\begin{array}{rl}
    & \argmin_{P^i_0,\ldots,P^i_N} \sum_{i=0}^M (1 - \lambda) L_i + \lambda \kappa_{\text{max}}^{i^2} \\\\
\text{s.t.:}
   & B_i(t) \subset \mathcal{P}_{\text{safe}}^i, \forall i, t\in[0,1] \\
   & B_i(t) \in C^1, \forall i \\
   & B_0(0) = P_{\text{start}}, \quad B_M(1) = P_{\text{end}}
\end{array}
\end{equation}
\noindent where:
\begin{itemize}
    \item $L_i$ is the arc length of curve segment $B_i$.
    \item $\kappa_{\text{max}}^{i^2}$ is the maximum squared curvature on segment $B_i$.
    \item $\lambda \in [0,1]$ is a user-defined parameter balancing path length and curvature minimization.
    \item Constraints ensure that each segment remains within corresponding safe polytopes ($\mathcal{P}_{\text{safe}}^i$),  maintains smoothness and continuity, and connects precisely from start to goal poses.
\end{itemize}

\section{Optimization formulation}
To practically solve problem \eqref{eq:prob_ideal}, we introduce a computationally efficient quadratic programming (QP) formulation using piece-wise quadratic Bezier (PWB) curves. The core idea is to balance smoothness, obstacle clearance, and computational efficiency within a single optimization framework. This section presents the detailed construction of constraints and the objective function.

\textbf{Construction of safe polytopes}. To explicitly incorporate safety margins and obstacle avoidance constraints, we propose a construction termed \emph{safe polytopes}. A safe polytope, denoted as $\mathcal{P}_{safe}$, is defined as a strict subset of the original polytope $\mathcal{P}$, enforcing a safety margin $\varepsilon$ from all obstacles and workspace boundaries. Mathematically, each safe polytope is described as follows:
$$\mathcal{P}_{safe} = \left\{ x \in \mathbb{R}^2 \mid H \cdot x \leq \overline{b}_{safe}\right\},$$

\noindent where $\overline{b}_{safe}$ is derived by shifting the original boundary vector  $\overline{b}$ inward by the safety margin $\varepsilon$.

\begin{figure}[h]
    \centering
    \includegraphics[width=0.7\linewidth]{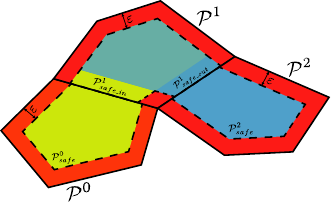}
    \caption{Illustration of connecting safe polytopes, facilitating smooth continuity between adjacent path segments.}
    \label{fig:safe_poly}
\end{figure}

To ensure smooth transitions between adjacent polytopes and maintain continuity, two distinct safe polytopes per original polytope are introduced:
$$\mathcal{P}_{safe\_in} = (H, \overline{b}_{safe\_in}), \quad \mathcal{P}_{safe\_out} = (H, \overline{b}_{safe\_out}).$$
These modified polytopes exclude the safety margin on their shared boundary, thus facilitating seamless connectivity, as illustrated in Fig.~\ref{fig:safe_poly}.

A safe polytope $\mathcal{P}_{safe}$ is defined as a subset $\mathcal{P}_{safe} \subset \mathcal{P}$, which includes a safety distance $\varepsilon$ for each inequality of the base polytope. Hence, the control points of each quadratic Bezier segment within polytope $i$ satisfy:
\begin{equation}\label{eq:safe}
\begin{array}{rl}
    HP^i_{0,1} &\le \overline{b}^i_{safe\_in} \\
    HP^i_{1,2} &\le \overline{b}^i_{safe\_out} 
\end{array}
\end{equation}
These linear inequality constraints ensure all Bezier curve segments remain entirely within safe, obstacle-free regions.

\textbf{Continuity and end-point constraints}. Smoothness and continuity between adjacent curve segments are enforced by ensuring both positional continuity $C^0$ and continuity of the first derivative ($C^1$) that translate into straightforward linear equalities for quadratic Bezier curves, expressed in terms of control points:
 \begin{equation}\label{eq:c01_cont}
 \begin{array}{rl}
    P^i_2 &= P^{i+1}_0 \\
    P^i_{2}-P^i_{1} &= P^{i+1}_1 - P^{i+1}_0
 \end{array}
 \end{equation}
\noindent where $P^i_2$ denotes the last control point of the polytope $i$, ensuring a smooth transition into the subsequent segment. Additionally, trajectory endpoints are defined by fixing the initial and final control points to given start and goal positions:
\begin{equation}
\begin{array}{rl} \label{eq:final}
        P^0_0 &= P_{start}\\
        P^M_2 &= P_{end} 
\end{array}
\end{equation}

\textbf{Objective Function Construction.} From the ideal objective in \eqref{eq:prob_ideal}, minimizing arc length and absolute curvature, is computationally expensive due to the nonlinearity of curvature. To improve efficiency, we adopt a quadratic formulation based on geometric properties of quadratic Bezier curves.
Empirical evidence shows that placing the intermediate control point $P^i_1$ near the midpoint of $P^i_0$ and $P^i_2$ effectively reduces curvature.
A formal mathematical justification of this property will be rigorously demonstrated in future work.

Consequently, the simplified objective function to be minimized for each segment $i$ is constructed as a quadratic form, explicitly balancing both curvature and path length:

\begin{equation}\label{eq:final_objective}
    J = \sum \limits_{i=0}^M \lVert P^i_1 - P^i_0 \rVert^2 + \lVert P^i_2 - P^i_1\rVert^2 +  \lambda \lVert P^i_2-P^i_0\rVert^2
\end{equation}

\noindent where the terms $\lVert P^i_1 - P^i_0 \rVert^2$ and $\lVert P^i_2 - P^i_1\rVert^2$ jointly approximate curvature reduction, encouraging smoothness, while the term $\lambda \lVert P^i_2-P^i_0\rVert^2$ explicitly encourages shorter paths. The user-defined parameter $\lambda > 0$ controls this trade-off, with lower values emphasizing smoothness  and higher values prioritizing shorter path length.

The resulting paths provide robust and smooth trajectories explicitly designed for safe robotic navigation within obstacle-cluttered environments, with suitable computational complexity for embedded, real-time implementations.

\section{Results and future work}
\textbf{Experimental setup}. To assess the performance and practical applicability of the proposed quadratic programming (QP) approach using piece-wise quadratic Bezier (PWB) curves, two representative scenarios were evaluated through simulations: a sparse obstacle environment and a narrow passage scenario. Our methodology was systematically compared against a previously proposed piece-wise linear (PWL) approach \cite{kloetzer2015optimizing}, considering both QP and nonlinear gradient-based optimization formulations. In total, four trajectory types were analyzed:

\begin{itemize}
\item PWL trajectories optimized via QP.
\item PWL trajectories optimized via nonlinear gradient.
\item PWB trajectories optimized via our proposed QP method.
\item PWB trajectories optimized via nonlinear gradient.
\end{itemize}

All simulations were conducted using MATLAB, employing the YALMIP toolbox and the \emph{quadprog} solver for QP formulations, and the gradient-based solver \emph{fmincon} for nonlinear optimization cases. The nonlinear PWL approach minimizes Euclidean distance, while our QP methods optimize squared distances for computational efficiency. Nonlinear Bezier optimization followed the original idealized curvature-length minimization objective (Eq. \eqref{eq:prob_ideal}), with minor adjustments for numerical stability.

The chosen parameters were: safety margin $\varepsilon = 0.2$, trade-off parameter for curvature and length $\lambda = 10$ for QP methods, and $\lambda = \frac{10}{11}$ for nonlinear optimization. Simulations were performed on a desktop computer with an AMD Ryzen 7 7800X3D processor (4.2 GHz) and 64 GB RAM.

\begin{figure*}[ht]
    \centering
    \begin{subfigure}[b]{0.49\textwidth}
        \centering
        \quad\includegraphics[width=0.81\linewidth]{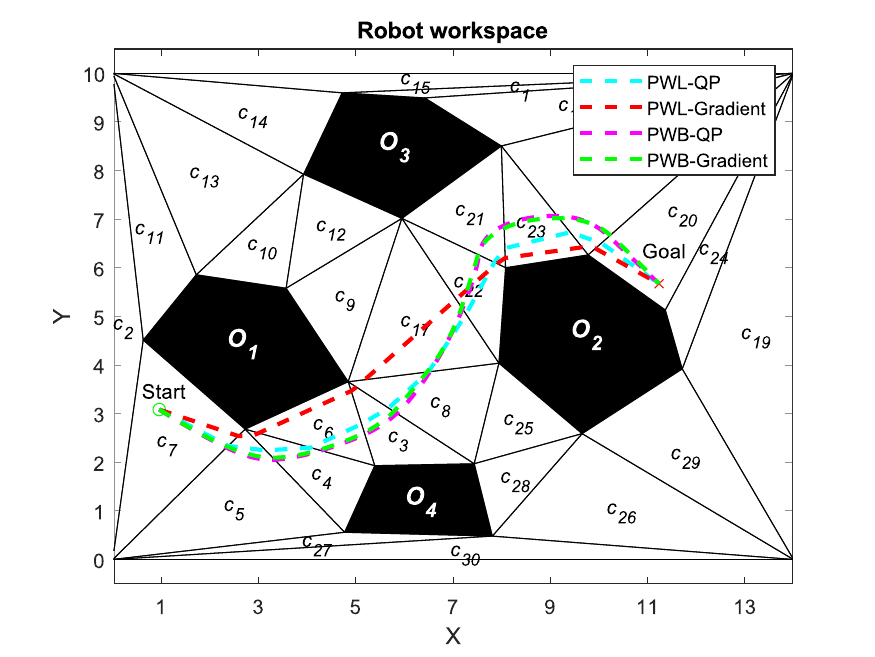}
        \caption{}
        \label{fig:sparse}
    \end{subfigure}
    \hfill
    \begin{subfigure}[b]{0.49\textwidth}
        \centering
        \includegraphics[width=0.81\linewidth]{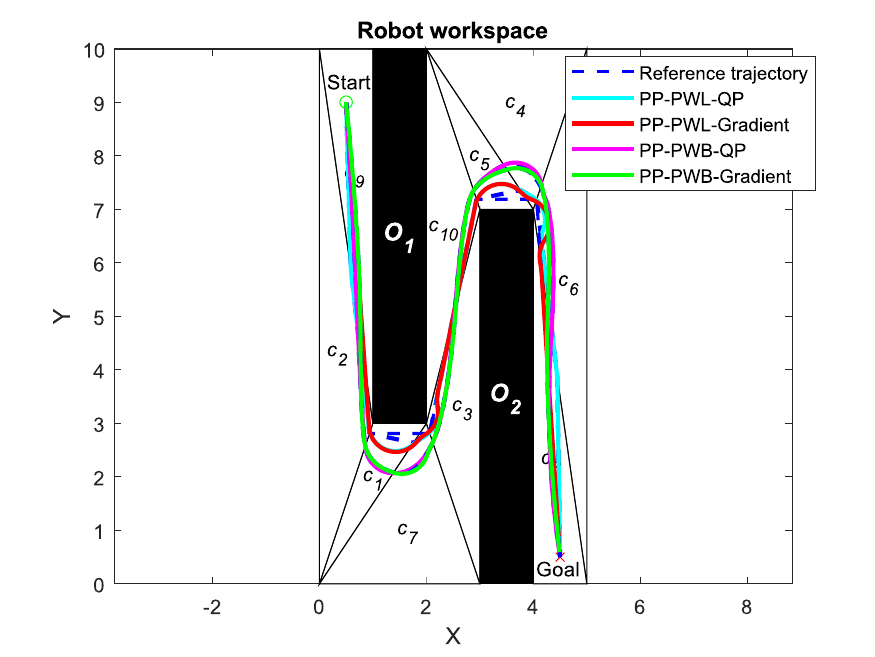}
        \caption{}
        \label{fig:narrow}
    \end{subfigure}
    \caption{Optimized trajectories for different environment scenarios.}
    \label{fig:comparison}
\end{figure*}

\textbf{Sparse environment scenario}. This environment represents a relatively open workspace populated with isolated convex polygonal obstacles. Fig.~\ref{fig:sparse} illustrates the resulting paths for all four tested optimization methods. As visible from the figure, all generated paths successfully maintained required safety margins and avoided obstacles. Table~\ref{tab:sparse} summarizes key quantitative metrics, including computational complexity, path length, and computation times. Although PWB trajectories involve higher computational complexity due to more decision variables and constraints, the total computation time remains well below 100 milliseconds, making the approach suitable for real-time applications. The trade-off, a minor increase of approximately 1–2 meters in path length, is compensated by significantly smoother and more robust trajectories compared to linear alternatives.

\begin{table}[h]
    \centering
    \caption{Performance comparison for sparse environment.}
    \begin{adjustbox}{width=0.95\columnwidth,center}
        \begin{tabular}{|c|cc|cc|}
            \hline
             \multirow{2}{*}{\diagbox{Metrics}{Trajectories}} & \multicolumn{2}{c|}{Piece-Wise Linear}    & \multicolumn{2}{c|}{Piece-Wise Bezier}    \\ \cline{2-5} 
             & \multicolumn{1}{c|}{QP} & Gradient & \multicolumn{1}{c|}{QP} & Gradient \\ \hline
             Computing Time [s]& \multicolumn{1}{c|}{\textbf{0.001}} & 0.029 & \multicolumn{1}{c|}{0.075} & 0.077\\ \hline
             Path Length [m]& \multicolumn{1}{c|}{12.71} & \textbf{11.79} & \multicolumn{1}{c|}{13.60} & 13.48\\ \hline
             Nr. of decision variables& \multicolumn{1}{c|}{\textbf{11}} & \textbf{11} & \multicolumn{1}{c|}{144} & 72\\ \hline
             Nr. of constraints& \multicolumn{1}{c|}{\textbf{22}} & \textbf{22} & \multicolumn{1}{c|}{336} & 192\\ \hline
        \end{tabular}
    \end{adjustbox}
    \label{tab:sparse}
\end{table}

\textbf{Narrow passage scenario}. This scenario was designed to challenge the trajectory-planning algorithm in constrained environments, where precise maneuvering is critical. Fig.~\ref{fig:narrow} displays the optimized paths obtained using the four trajectory generation methods. Table~\ref{tab:narrow} summarizes quantitative metrics from this scenario, including path execution times, maximum trajectory deviation, and maximum curvature, evaluated by simulating an Ackermann-steered mobile robot controlled using a standard Pure-Pursuit controller from the RMTool toolbox~\cite{IPGoMaKl15}. Simulation parameters included a maximum linear velocity of $1 m/s$, maximum steering angle of $35^o$, and a look-ahead horizon of $15$ steps. Notably, PWB trajectories exhibited substantially lower deviation from planned paths compared to PWL trajectories, significantly enhancing safety and precision in constrained spaces. Even at maximum allowable curvature, the deviations remained within the defined safety margins. This indicates improved practical reliability and robustness of PWB trajectories under realistic robotic conditions.

\begin{table}[h]
    \centering
    \caption{Performance comparison for narrow passage.}
    \begin{adjustbox}{width=0.95\columnwidth,center}
        \begin{tabular}{|c|cc|cc|}
            \hline
             \multirow{2}{*}{\diagbox{Metrics}{Trajectories}} & \multicolumn{2}{c|}{Piece-Wise Linear}    & \multicolumn{2}{c|}{Piece-Wise Bezier}    \\ \cline{2-5} 
             & \multicolumn{1}{c|}{QP} & Gradient & \multicolumn{1}{c|}{QP} & Gradient \\ \hline
             Computing Time [s]& \multicolumn{1}{c|}{\textbf{8.52e-4}} & 0.019 & \multicolumn{1}{c|}{0.037} & 0.1086\\ \hline
             Path Length [m]& \multicolumn{1}{c|}{19.77} & \textbf{19.63} & \multicolumn{1}{c|}{21.62} & 21.50\\ \hline
             Execution Time [s]& \multicolumn{1}{c|}{19.92} & \textbf{19.77} & \multicolumn{1}{c|}{21.61} & 21.46\\ \hline
             Maximum Trajectory Deviation [m]& \multicolumn{1}{c|}{0.232} & 0.343 & \multicolumn{1}{c|}{0.073} & \textbf{0.034}\\ \hline
             Maximum Curvature [rad/m]& \multicolumn{1}{c|}{2.0} & 2.0 & \multicolumn{1}{c|}{2.0} & 2.0\\ \hline
        \end{tabular}
    \end{adjustbox}
    \label{tab:narrow}
\end{table}
\textbf{Discussion}. Results demonstrate the practical benefits and feasibility of the proposed quadratic programming approach using quadratic Bezier curves for robotic path planning. Despite moderately increased computational complexity compared to linear methods, our QP-based PWB method maintains computational efficiency compatible with real-time robotics applications. The marginal increase in computational time and trajectory length is well-compensated by considerable improvements in smoothness, precision, and robustness. Moreover, nonlinear optimization methods offer minor performance advantages at significantly higher computational cost, highlighting the suitability of our quadratic programming approach as an excellent balance between computational efficiency and trajectory quality.

\textbf{Future work}. Future research will focus on further optimizing the proposed QP formulation, particularly reducing computational complexity through improved constraint formulations and exploring formal proofs of geometric properties utilized in our approach. Additionally, real-world experimental validation will be conducted on robotic platforms with explicit kinematic and dynamic constraints, confirming the practical applicability of the proposed methodology.

\bibliographystyle{ieeetr}
\bibliography{bib}

\end{document}